\documentclass[]{article}
\usepackage{proceed2e}

\usepackage{times}
\usepackage{graphicx}
\usepackage{amsmath}
\usepackage{amssymb}
\usepackage{paralist}
\usepackage{color}
\usepackage{url}
\usepackage[round,sort&compress]{natbib}

\newcommand{\defoccur}[1]{\textsl{#1}}

\newcommand{\LincolnLogs}{\textsc{Lincoln Logs}}
\newcommand{\LincolnLog}{\textsc{Lincoln Log}}

\newcommand{\gPb}{\textsc{gPb}}

\DeclareMathOperator*{\argmax}{argmax}

\newcommand{\true}{\mathbf{true}}
\newcommand{\false}{\mathbf{false}}

\title{Seeing Unseeability to See the Unseeable}

\author{Siddharth Narayanaswamy, Andrei Barbu, and Jeffrey Mark Siskind}
\date{Purdue University\\
School of Electrical and Computer Engineering\\
465 Northwester Avenue\\
West Lafayette IN 47907-2035 USA\\
{\tt\small andrei@0xab.com,\{snarayan,qobi\}@purdue.edu}}

\begin{document}

\maketitle

\let\thefootnote\relax\footnotetext{Additional images and videos as well as all
  code and datasets are available at
  \url{http://engineering.purdue.edu/~qobi/arxiv2012c}.}

\begin{abstract}
We present a framework that allows an observer to determine occluded portions
of a structure by finding the maximum-likelihood estimate of those occluded
portions consistent with visible image evidence and a consistency model.
Doing this requires determining which portions of the structure are occluded in
the first place.
Since each process relies on the other, we determine a solution to both
problems in tandem.
We extend our framework to determine confidence of one's assessment of which
portions of an observed structure are occluded, and the estimate of that
occluded structure, by determining the sensitivity of one's assessment to
potential new observations.
We further extend our framework to determine a robotic action whose execution
would allow a new observation that would maximally increase one's confidence.
\end{abstract}

\section{Introduction}
\label{sec:introduction}

\begin{quote}
[T]here are known knowns; there are things we know we know.
We also know there are known unknowns; that is to say we know there are some
things we do not know.
But there are also unknown unknowns--–the ones we don't know we don't know.
\begin{flushright}
Donald Rumsfeld (12 February 2002)
\end{flushright}
\end{quote}

People exhibit the uncanny ability to see the unseeable.
The colloquial exhortation \emph{You have eyes in the back of your head!}
expresses the assessment that someone is making correct judgements as if they
could see what is behind them, but obviously cannot.
People regularly determine the properties of occluded portions of objects from
observations of visible portions of those objects using general world knowledge
about the consistency of object properties.
Psychologists have demonstrated that the world knowledge that can influence
perception can be high level, abstract, and symbolic, and not just related to
low-level image properties such as object class, shape, color, motion, and
texture.
For example, \cite{FreydPC88} showed that physical forces, such as
gravity, and whether such forces are in equilibrium, due to support and
attachment relations, influences visual perception of object location in adults.
\cite{Baillargeon86, Baillargeon87} showed that knowledge of
substantiality, the fact that solid objects cannot interpenetrate, influences
visual object perception in young infants.
\cite{Streri1988} showed that knowledge about object rigidity
influences both visual and haptic perception of those objects in young infants.
Moreover, such influence is cross modal: observable haptic perception
influences visual perception of unobservable properties and observable visual
perception influences haptic perception of unobservable properties.
\cite{Wynn1998} showed that material properties of objects, such as
whether they are countable or mass substances, along with abstract properties,
such as the number of countable objects and the quantity of mass substances, and
how they are transferred between containers, influences visual perception in
young infants.
Similar results exist for many physical properties such as relative mass,
momentum, etc.
These results demonstrate that people can easily integrate information from
multiple sources together with world knowledge to see the unseeable.

People so regularly invoke the ability to see the unseeable that we often don't
realize that we do so.
If you observe a person entering the front door of a house and later see them
appear from behind the house without seeing them exit a door, you easily
see the unseeable and conclude that there must be an unseen door to the house.
But if one later opens the garage door or the curtain covering a large
living-room bay window in the front of the house so that you see through the
house and see the back door you no longer need to invoke the ability to see
the unseeable.
A more subtle question then arises: when must you invoke the ability to see the
unseeable?
In other words how can you see unseeability, the inability to see?
This question becomes particularly thorny since, as we will see, it can
involve a chicken-and-egg problem: seeing the unseen can require seeing the
unseeability of the unseen and seeing the unseeability of the unseen can
require seeing the unseen.

The ability to see unseeability and to see the unseeable can further
dramatically influence human behavior.
We regularly and unconsciously move our heads and use our hands to open
containers to render seeable what was previously unseeable.
To realize that we need to do so in the first place, we must first see the
unseeability of what we can't see.
Then we must determine how to best use our collective perceptual, motor, and
reasoning affordances to remedy the perceptual deficiency.

We present a general computational framework for seeing unseeability to see the
unseeable.
We formulate and evaluate a particular instantiation of this general framework
in the context of a restricted domain, namely \LincolnLogs, a children's
assembly toy where one constructs assemblies from a small inventory of
component logs.
The two relevant aspects of this domain that facilitate its use for
investigating our general computational framework are (a)~that
\LincolnLog\ assemblies suffer from massive occlusion and (b)~that a simple but
rich expression of world knowledge, in the form of constraints on valid
assemblies, can mitigate the effects of such occlusion.

\begin{figure}
\begin{center}
\begin{tabular}{@{}c@{\hspace*{2pt}}c@{}}
\includegraphics[width=0.23\textwidth]{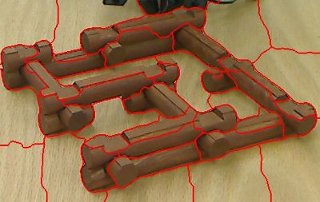}&
\includegraphics[width=0.23\textwidth]{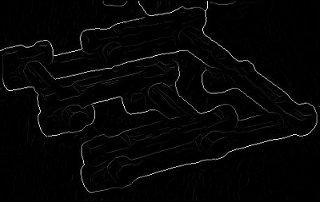}\\
(a)&(b)
\end{tabular}
\end{center}
\vspace{-2ex}
\caption{(a)~A state-of-the-art segmentation method, Normalized Cut
  \citep{Shi2000}, does not segment out the log parts.
(b)~A state-of-the-art edge detector, \gPb\ \citep{Maire2008}, does not reliably
find edges separating adjacent logs or log ends.}
\label{fig:stateoftheart}
\vspace{-2ex}
\end{figure}

While \LincolnLogs\ are a children's toy, this domain is far from a toy when it
comes to computer vision.
The task of structure estimation, determining, from an image, the correct
combination of component logs used to construct an assembly and how they are
combined, is well beyond state-of-the-art methods in computer vision.
We have not found any general-purpose image segmentation methods that can
determine the image boundaries of the visible component logs
(see Fig.~\ref{fig:stateoftheart}a).
Moreover, the uniform matte color and texture of the logs, together with the
fact that logs are placed in close proximity and the fact that the majority of
any structure is in self shadow means every edge-detection method that we have
tried fails to find the boundaries between adjacent logs (see
Fig.~\ref{fig:stateoftheart}b).
This is even before one considers occlusion, which only makes matters worse.

Not only is the computer-vision problem for this domain immensely difficult,
the computational problem is rich as well.
We present methods for seeing the unseeable (in section~\ref{sec:structure})
and seeing unseeability (in section~\ref{sec:visibility}) based on precise
computation of the maximum-likelihood structure estimate conditioned on world
knowledge that marginalizes over image evidence.
We further present (in section~\ref{sec:confidence}) a rational basis for
determining confidence in one's structure estimate despite unseeability based
on precise computation of the amount of evidence needed to override a uniform
prior on the unseeable.
And we finally present (in section~\ref{sec:integration}) an active-vision
decision-making process for determining rational behavior in the presence of
unseeability based on precise computation of which of several available
perception-enhancing actions one should take to maximally improve the
confidence in one's structure estimate.
We offer experimental evaluation of each of these methods in
section~\ref{sec:results}, compare against related work in
section~\ref{sec:related}, and conclude with a discussion of potential
extensions in section~\ref{sec:conclusion}.

\section{Structure Estimation}
\label{sec:structure}

In previous work \cite{Siddharth2011} presented an approach for using a visual
language model for improving recognition accuracy on compositional visual
structures in a generative visual domain, over the raw recognition rate of the
part detectors---by analogy to the way speech recognizers use a human language
model to improve recognition accuracy on utterances in a generative linguistic
domain, over the raw recognition rate of the phoneme detectors.
In this approach, a complex object is constructed out of a collection of parts
taken from a small part inventory.
A language model, in the form of a stochastic constraint-satisfaction problem
(CSP) \citep{Lauriere1978}, characterizes the constrained way object parts can
combine to yield a whole object and significantly improves the recognition rate
of the whole structure over the infinitesimally small recognition rate that
would result from unconstrained application of the unreliable part detectors.
Unlike the speech-recognition domain, where (except for coarticulation) there
is acoustic evidence for all phonemes, in the visual domain there may be
components with no image evidence due to occlusion.
A novel aspect of applying a language model in the visual domain instead of the
linguistic domain is that the language model can additionally help in
recovering occluded information.

This approach was demonstrated in the domain of \LincolnLogs, a children's
assembly toy with a small part inventory, namely, 1-notch, 2-notch, and 3-notch
logs, whose CAD models are provided to the system.
In this domain, a grammatical \LincolnLog\ structure contains logs that are
parallel to the work surface and organized on alternating layers oriented in
orthogonal directions.
Logs on each layer are mutually parallel with even spacing, thereby imposing a
symbolic grid on the \LincolnLog\ assembly.
The symbolic grid positions $q=(i,j,k)$ refer to points along log medial axes
at notch centers.
One can determine the camera-relative pose of this symbolic grid without any
knowledge of the assembly structure by fitting the pose to the two predominant
directions of image edges that result from the projection of the logs to the
image plane.

Each grid position may be either unoccupied, denoted by~$\emptyset$, or
occupied with the~$n^{\textrm{th}}$ notch, counting from zero, of a log
with~$m$ notches, denoted by $(m,n)$.
Estimating the structure of an assembly reduces to determining the occupancy at
each grid position, one of the seven possibilities: $\emptyset$, $(1,0)$,
$(2,0)$, $(2,1)$, $(3,0)$, $(3,1)$, and $(3,2)$.
This is done by constructing a discrete random variable~$Z_q$ for each grid
position~$q$ that ranges over these seven possibilities, mutually constraining
these random variables together with other random variables that characterize
the image evidence for the component logs using the language model, and finding
a maximum-likelihood consistent estimate to the random variables~$Z_q$.

\begin{figure}
\begin{center}
\includegraphics[width=0.48\textwidth]{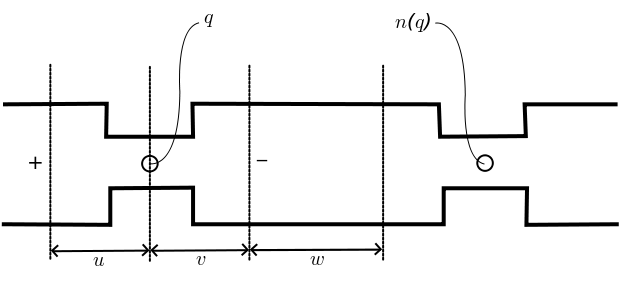}
\vspace{-3ex}
\end{center}
\caption{The random variables~$Z^+_q$ and~$Z^-_q$ that correspond to log
  ends for grid position~$q$ and the random variables~$Z^u_q$, $Z^v_q$,
  and~$Z^w_q$ that correspond to log segments.}
\vspace{-2ex}
\label{fig:endsandsegs}
\end{figure}

Several forms of image evidence are considered for the component logs.
\LincolnLogs, being cylindrical parts, generate two predominant image (log)
features: ellipses that result from the perspective projection of circular log
ends and line segments that result from the perspective projection of
cylindrical walls.
The former are referred to as \defoccur{log ends} and the latter as
\defoccur{log segments}.
Log ends can potentially appear only at a fixed distance on either side of a
grid position.
Boolean random variables~$Z^+_q$ and~$Z^-_q$ are constructed to encode the
presence or absence of a log end at such positions.
There are two kinds of log segments: those corresponding to the portion of a
log between two notches and those corresponding to the portions of a log end
that extend in front of or behind the two most extreme notches.
Given this, three Boolean random variables~$Z^u_q$, $Z^v_q$, and~$Z^w_q$ are
constructed for each grid position~$q$ that encode the presence or absence of
such log segments for the bottoms of logs, i.e. log segments between a grid
position and the adjacent grid position below.
Fig.~\ref{fig:endsandsegs} depicts the log ends and log segments that
correspond to a given grid position as described above.

A stochastic CSP encodes the validity of an assembly.
Image evidence imposes priors on the random variables~$Z^+_q$,
$Z^-_q$, $Z^u_q$, $Z^v_q$, and~$Z^w_q$ and structure estimation is performed by
finding a maximum-likelihood solution to this stochastic CSP.\@
When formulating the constraints, the adjacent grid position below~$q$ is
referred to as~$b(q)$ and the adjacent grid position further from the origin
along the direction of the grid lines for the layer of~$q$ is referred to
as~$n(q)$.
Ignoring boundary conditions at the perimeter of the grid, the grammar of
\LincolnLogs\ can be formulated as the following constraints:
\begin{compactenum}[a)]
\item 2-notch logs occupy two adjacent grid points
\label{constraintA}
\item 3-notch logs occupy three adjacent grid points
\label{constraintB}
\item 1- and 2-notch logs must be supported at all notches
\label{constraintC}
\item 3-notch logs must be supported in at least 2 notches
\label{constraintD}
\item log ends must be at the ends of logs
\label{constraintE}
\item short log segments indicate occupancy above or below
\label{constraintF}
\item long log segments indicate presence of a multi-notch log above or below
\label{constraintG}
\end{compactenum}
Boundary conditions are handled by stipulating that the grid positions beyond
the perimeter are unoccupied, enforcing the support requirement (constraints
\ref{constraintC}--\ref{constraintD}) only at layers above the lowest layer,
and enforcing log-segment constraints (\ref{constraintF}--\ref{constraintG}) for
the layer above the top of the structure.
Structure estimation is performed by first establishing priors over the random
variables~$Z^+_q$, $Z^-_q$, $Z^u_q$, $Z^v_q$, and~$Z^w_q$ that correspond to
log features using image evidence and establishing a uniform prior over the
random variables~$Z_q$ that correspond to the latent structure.
This induces a probability distribution over the joint support of these random
variables.
The random variables that correspond to log features are marginalized and the
resulting marginal distribution is conditioned on the language model~$\Phi$.
Finally, the assignment to the collection, $\mathbf{Z}$, of random
variables~$Z_q$, that maximizes this conditional marginal probability is
computed.
\begin{displaymath}
\argmax_{\mathbf{Z}}\hspace{-13pt}
\sum_{\substack{\mathbf{Z}^+,\mathbf{Z}^-,\mathbf{Z}^u,\mathbf{Z}^v,\mathbf{Z}^w\\
    \Phi\left[\mathbf{Z},\mathbf{Z}^+,\mathbf{Z}^-,\mathbf{Z}^u,\mathbf{Z}^v,\mathbf{Z}^w\right]}}
\hspace{-10pt}\Pr\left(\mathbf{Z},\mathbf{Z}^{+},\mathbf{Z}^{-},\mathbf{Z}^u,\mathbf{Z}^v,\mathbf{Z}^w\right)
\end{displaymath}
While, in principle, this method can determine the conditional probability
distribution over consistent structures given image evidence, doing so is
combinatorially intractable.
The conditional marginalization process is made tractable by pruning
assignments to the random variables that violate the grammar~$\Phi$ using arc
consistency \citep{Mackworth1977}.
The maximization process is made tractable by using a branch-and-bound algorithm
\citep{Land1960} that maintains upper and lower bounds on the maximal
conditional marginal probability.
Thus instead of determining the distribution over structures, this yields a
single most-likely consistent structure given the image evidence, along with
its probability.

\section{Visibility Estimation}
\label{sec:visibility}

Image evidence for the presence or absence of each log feature is obtained
independently.
Each log feature corresponds to a unique local image property when projected to
the image plane under the known camera-relative pose.
A prior over the random variable associated with a specific log feature can be
determined with a detector that is focused on the expected location and shape
of that feature in the image given the projection.
This assumes that the specific log feature is visible in the image, and not
occluded by portions of the structure between the camera and that log feature.
When the log feature $f$, a member of the set $\{+,-,u,v,w\}$ of the five
feature classes defined above, at a position $q$, is not visible, the prior can
be taken as uniform, allowing the constraints in the grammar to fill in unknown
information.
We represent the visibility of a feature by the boolean variable $V_q^f$.
\begin{equation*}
\begin{array}{ll}
\Pr(Z_q^{\,f}=\true)\propto\text{image evidence}&\text{when $V_q^{\,f}=\true$}\\
\Pr(Z_q^{\,f}=\false)=\frac{1}{2}&\text{otherwise}
\end{array}
\end{equation*}
In order to do so, it is necessary to know which log features are visible and
which are occluded so that image evidence is only applied to construct a prior
on visible log features and a uniform prior is constructed for occluded log
features.
Thus, in Rumsfeld's terminology, one needs to know the known unknowns in order
to determine the unknowns.
This creates a chicken-and-egg problem.
To determine whether a particular log feature is visible, one must know the
composition of the structure between that feature and the camera and, to
determine the structure composition, one must know which log features are
visible.
While earlier \cite{Siddharth2011} demonstrated successful automatic
determination of log occupancy at occluded log positions, we could only do so
given manual annotation of log-feature visibility.
In other words, while earlier we were able to automatically infer $Z_q$, it
required manual annotation of $V_q^f$.
Further, determining $V_q^f$ required knowledge of $Z_q$.

We extend this prior work to automatically determine
visibility of log features in tandem with log occupancy.
Our novel contribution in this section is mutual automatic determination of
both $Z_q$ and $V_q^f$.
We solve the chicken-and-egg problem inherent in doing so with an iterative
algorithm reminiscent of expectation maximization (EM) \citep{Baum72, BaumPSW70,
  DempsterLR77}.
We start with an initial estimate of the visibility of each log feature.
We then apply the structure estimation procedure developed in 
\cite{Siddharth2011} to estimate the occupancy of each symbolic grid position.
We then use the estimated structure to recompute a new estimate of log-feature
visibility, and iterate this process until a fixpoint is reached.
There are two crucial components of this process: determining the initial
log-feature visibility estimate and reestimating log-feature visibility from an
estimate of structure.

We determine the initial log-feature visibility estimate (i.e. $V_q^f$) by
assuming that the structure is a rectangular prism whose top face and two
camera-facing front faces are completely visible.
In this initial estimate, log features on these three faces are visible and
log features elsewhere are not.
We use the camera-relative pose of the symbolic grid (which can be determined
without any knowledge of the structure) together with maximal extent of each of
the three symbolic grid axes (i.e., three small integers which are currently
specified manually) to determine the visible faces.
This is done as follows.
We determine the image positions for four corners of the base of this
rectangular prism: the origin~$(0,0,0)$ of the symbolic grid, the two extreme
points~$(i_{\textrm{max}},0,0)$ and~$(0,0,k_{\textrm{max}})$ of the two
horizontal axes in the symbolic grid, and the symbolic grid
point~$(i_{\textrm{max}},0,k_{\textrm{max}})$.
We select the bottommost three such image positions as they correspond to the
endpoints of the lower edges of the two frontal faces.
It is possible, however, that one of these faces is (nearly) parallel to the
camera axis and thus invisible.
We determine that this is the case when the angle subtended by the two lower
edges previously determined is less than 110$^{\circ}$ and discard the face
whose lower edge has minimal image width.

We update the log-feature visibility estimate from a structure estimate by
rendering the structure in the context of the known camera-relative pose of the
symbolic grid.
When rendering the structure, we approximate each log as the bounding cylinder
of its CAD model.
We characterize each log feature with a fixed number of points, equally spaced
around circular log ends or along linear log segments and trace a ray from each
such point's 3D position to the camera center, asking whether that ray
intersects some bounding cylinder for a log in the estimated structure.
We take a log feature to be occluded when 60\% or more of such rays intersect
logs in the estimated structure.
Our method is largely insensitive to the particular value of this threshold.
It only must be sufficiently low to label log features as invisible when they
actually are invisible.
Structure estimation is not adversely affected by a moderate number of log
features that are incorrectly labeled as invisible when they are actually
visible because it can use the grammar to determine occupancy of grid positions
that correspond to such log features.

We can perform such rendering efficiently by rasterization.
For each log feature, we begin with an empty bitmap.
We iterate over each log feature and each occupied grid position that lies
between that log feature and the camera center and render a projection of the
bounding cylinder of the log at that grid position on the bitmap.
This renders all possible occluders for each log feature allowing one to
determine visibility by counting the rendered pixels at points in the bitmap
that correspond to the projected rays.

The above process might not reach a fixpoint and instead may enter a finite
loop of pairs of visibility and structure estimates.
In practice, this process either reaches a fixpoint within three to four
iterations or enters a loop of length two within three to four iterations,
making loop detection straightforward.
When a loop is detected, we select the structure in the loop with the highest
probability estimate.

\section{Structure-Estimation Confidence}
\label{sec:confidence}

While the structure estimation process presented by \cite{Siddharth2011} can
determine the occupancy of a small number of grid positions when only a single
set of occupancy values is consistent with the grammar and the image evidence,
it is not clairvoyant; it cannot determine the structure of an assembly when a
large part of that assembly is occluded and many different possible structures
are consistent with the image evidence.
In this case, we again have an issue of unknowns vs. known unknowns: how can
one determine one's confidence in one's structure estimation.
If we could determine the conditional distribution over consistent structures
given image evidence, $P(\mathbf{Z}|I)$, we could take the entropy of this
distribution, $H(\mathbf{Z}|I)$, as a measure of confidence.
However, as discussed previously, it is intractable to compute this distribution
and further intractable to compute its entropy.
Thus we adopt an alternate means of measuring confidence in the result of the
structure-estimation process.

Given a visibility estimate, $V_q^f$, a structure estimate, $\mathbf{Z}$, and
the priors on the random variables associated with log features computed with
image evidence, $Z_q^f$, one can marginalize over the random variables
associated with visible log features and compute the maximum-likelihood
assignment to the random variables associated with occluded log features,
$\hat{\mathbf{Z}}^{\,f}$, that is consistent with a given structure estimate.
\begin{equation*}
\hat{\mathbf{Z}}^{\,f}=
\argmax_{\substack{Z_q^{\,f}\\V_q^{\,f}=\false}}\hspace{-4.5ex}
\sum_{\substack{Z_q^{\,f}\\V_q^{\,f}=\true\\
\Phi[\mathbf{Z},\mathbf{Z}^{+},\mathbf{Z}^{-},\mathbf{Z}^u,\mathbf{Z}^v,\mathbf{Z}^w]}}\hspace{-6ex}
\Pr(\mathbf{Z},\mathbf{Z}^{+},\mathbf{Z}^{-},\mathbf{Z}^u,\mathbf{Z}^v,\mathbf{Z}^w)
\end{equation*}
One can then ask the following question: what is the maximal amount~$\delta$
that one can shift the probability mass on the random variables associated
with occluded log features \emph{away} from the uniform prior, reassigning
that shifted probability mass to the opposite element of the support of that
random variable from the above maximum-likelihood assignment, such that
structure estimation yields the same estimated structure.
Or in simpler terms,
\begin{quote}
\emph{How much hypothetical evidence of occluded log features is needed to
  cause me to change my mind away from the estimate derived from a uniform
  prior on such occluded features?}
\end{quote}
We compute this $\delta$ using a modified structure estimation step
\begin{equation*}
\argmax_{\mathbf{Z}}\hspace{-5ex}
\sum_{\substack{\mathbf{Z}^{+},\mathbf{Z}^{-},\mathbf{Z}^u,\mathbf{Z}^v,\mathbf{Z}^w\\
\Phi[\mathbf{Z},\mathbf{Z}^{+},\mathbf{Z}^{-},\mathbf{Z}^u,\mathbf{Z}^v,\mathbf{Z}^w]}}\hspace{-5ex}
\Pr(\mathbf{Z},\mathbf{Z}^{+},\mathbf{Z}^{-},\mathbf{Z}^u,\mathbf{Z}^v\,\mathbf{Z}^w)=\mathbf{Z}
\end{equation*}
when, for all $q^{\,f}$ where $V_q^{\,f}=\false$
\begin{equation*}
\begin{array}{l}
\Pr(Z_q^{\,f}=\neg\hat{Z}_q^{\,f})=\frac{1}{2}+\delta\\
\Pr(Z_q^{\,f}=\hat{Z}_q^{\,f})=\frac{1}{2}-\delta
\end{array}
\end{equation*}
We call such a~$\delta$ the \defoccur{estimation tolerance}.
Then, for any estimated structure, one can make a confidence judgment by
comparing the estimation tolerance to an overall tolerance threshold~$\delta^*$.
One wishes to select a value for~$\delta^*$ that appropriately trades off false
positives and false negatives in such confidence judgements: we want to
minimize the cases that result in a positive confidence assessment for an
incorrect structure estimate and also minimize the cases that result in a
negative confidence assessment for a correct structure estimate.
Because the methods we present in the next section can gather additional
evidence in light of negative confidence assessment in structure estimation,
the former are more hazardous than the latter because the former preclude
gathering such additional evidence and lead to an ultimate incorrect structure
estimate while the latter simply incur the cost of such additional evidence
gathering.
Because of this asymmetry, our method is largely insensitive to the particular
value of~$\delta^*$ so long as it is sufficiently high to not yield excessive
false positives.
We have determined empirically that setting $\delta^*=0.2$ yields a good
tradeoff: only~3/105 false positives and~7/105 false negatives on our corpus.

One can determine the estimation tolerance by binary search for the smallest
value of $\delta\in(0,0.5)$ that results in a different estimated structure.
However, this process is time consuming.
But we don't actually need the value of~$\delta$; we only need to determine
whether $\delta<\delta^*$.
One can do this by simply asking whether the estimated structure, $\mathbf{Z}$,
changes when the probabilities are shifted by~$\delta^*$
\begin{equation*}
\begin{array}{l}
\Pr(Z_q^{\,f}=\neg\hat{Z}_q^{\,f})=\frac{1}{2}+\delta^{*}\\
\Pr(Z_q^{\,f}=\hat{Z}_q^{\,f})=\frac{1}{2}-\delta^{*}
\end{array}
\end{equation*}
This involves only a single new structure estimation.
One can make this process even faster by initializing the branch-and-bound
structure-estimation algorithm with the probability of the original structure
estimate given the modified distributions for the random variables associated
with occluded log features.

\section{Gathering Additional Evidence to Improve Structure Estimation}
\label{sec:integration}

\begin{figure}
\begin{center}
\includegraphics[width=0.47\textwidth]{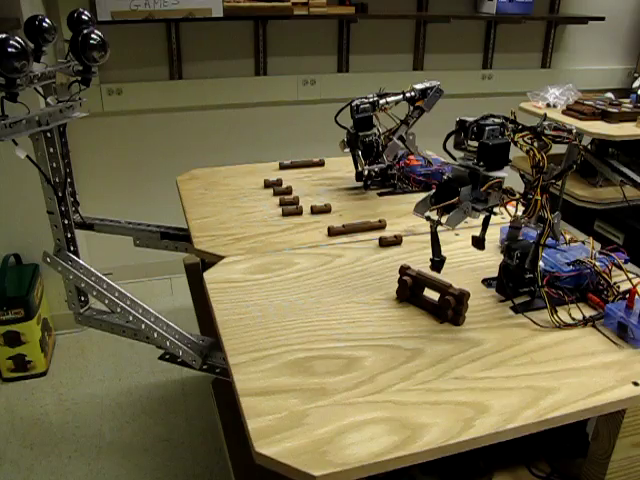}
\end{center}
\caption{Our novel robotic environment for performing structure estimation.
Note that the head can rotate 180$^{\circ}$ about the workspace, under computer
control, to image the assembly from different viewpoints, and the robot arm can
disassemble the structure on the workspace.}
\label{fig:robot}
\end{figure}

Structure estimation can be made more reliable by integrating multiple sources
of image evidence.
We perform structure estimation in a novel robotic environment, illustrated in
Fig.~\ref{fig:robot}, that facilities automatically gathering multiple sources
of image evidence as needed.
The structures are assembled in the robot workspace.
This workspace is imaged by a camera mounted on a pendulum arm that can rotate
180$^{\circ}$ about the workspace, under computer control, to image the
assembly from different viewpoints.
This can be used to view portions of the assembly that would otherwise be
occluded.
Moreover, a robotic arm can disassemble a structure on the workspace.
This can be used to reveal the lower layers of a structure that would
otherwise be occluded by higher layers.
These methods can further be combined.
Generally speaking, we seek a method for constraining a single estimate
of an initial structure with multiple log features derived from different
viewpoints and different stages of disassembly.

We can do this as follows.
Let~$\mathbf{Z}$ be a collection of random variables~$Z_q$ associated with log
occupancy for a given initial structure.
Given multiple views $i=1,\ldots,n$ with collections~$\mathbf{Z}_i$ of random
variables~$Z^+_q$, $Z^-_q$, $Z^u_q$, $Z^v_q$, and~$Z^w_q$ associated with the
image evidence for log features from those views, we can compute
\begin{displaymath}
\argmax_{\mathbf{Z}}\hspace{-12pt}
\sum_{\substack{\mathbf{Z}_1\ldots\mathbf{Z}_n\\
    \Phi\left[\mathbf{Z},\mathbf{Z}_1\right]\wedge\ldots\wedge\left[\mathbf{Z},\mathbf{Z}_n\right]}}
\hspace{-10pt}\Pr\left(\mathbf{Z},\mathbf{Z}_1,\ldots,\mathbf{Z}_n\right)
\end{displaymath}
Only two issues arise in doing this.
First, we do not know the relative camera angles of the different views.
Even though one can estimate the camera-relative pose of the structure
independently for each view, this does not yield the registration between these
views.
There are only four possible symbolic orientations of the structure in each
view so for~$n$ views we need only consider~$4^{n-1}$ possible combinations of
such symbolic orientations.
We can search for the combination that yields the maximum-likelihood structure
estimate.
We do this search greedily, incrementally adding views to the
structure-estimation process and registering each added view by searching for
the best among the four possible registrations.
Second, in the case of partial disassembly, we need to handle the fact that the
partially disassembled structure is a proper subset of the initial structure.
We do this simply by omitting random variables associated with log features
for logs that are known to have been removed in the disassembly process and not
instantiating constraints that mention such omitted random variables.

We can combine the techniques from section~\ref{sec:confidence} with these
techniques to yield an active-vision \citep{Bajcsy1988} approach to producing a
confident and correct structure estimate.
One can perform structure estimation on an initial image and assess one's
confidence in that estimate.
If one is not confident, one can plan a new observation, entailing either a new
viewpoint, a partial-disassembly operation, or a combination of the two and
repeat this process until one is sufficiently confident in the estimated
structure.
Only one issue arises in doing this.
One must plan the new observation.
We do so by asking the following question:
\begin{quote}
\emph{Which of the available actions maximally increases confidence?}
\end{quote}
Like before, if we could determine the conditional distribution over
consistent structures given image evidence, we could compute the decrease in
entropy that each available action would yield and select the action that
maximally decreases entropy.
But again, it is intractable to compute this distribution and further
intractable to compute its entropy.
Thus we adopt an alternate means of measuring increase in confidence.

Given visibility estimates $V_{iq}^f$ for view $i$ of the $n$ current views
along with a structure estimate $\mathbf{Z}$ constructed from those views, and
priors on the random variables associated with log features computed with image
evidence for each of these views $Z_{iq}^f$, one can marginalize over the
random variables associated with visible log features, $V_{iq}^f = \true$, and
compute the maximum-likelihood assignment $\hat{\mathbf{Z}}^{\,f}$ to the
random variables associated with occluded log features that is consistent with
a given structure estimate:
\begin{equation*}
\hat{\mathbf{Z}}^{\,f}=\argmax_{\substack{Z_{iq}^{\,f}\\V_{iq}^{\,f}=\false}}
\sum_{\substack{Z_{iq}^{\,f}\\V_{iq}^{\,f}=\true\\
\Phi[\mathbf{Z},\mathbf{Z}_1]\wedge\ldots\wedge\Phi[\mathbf{Z},\mathbf{Z}_n]}}
\Pr(\mathbf{Z},\mathbf{Z}_1,\ldots,\mathbf{Z}_n)
\end{equation*}
We can further determine those log features that are invisible in all current
views but visible in a new view $j$ that would result from a hypothetical
action under consideration.
One can then ask the following question: what is the maximal amount~$\delta'$
that one can shift the probability mass on these random variables \emph{away}
from the uniform prior, reassigning that shifted probability mass to the
opposite element of the support of that random variable from the above
maximum-likelihood assignment, such that structure estimation when adding the
new view yields the same estimated structure.
Or in simpler terms,
\begin{quote}
\emph{For a given hypothetical action, how much hypothetical evidence of
  log features that are occluded in all current views is needed in an imagined
  view resulting from that action where those log features are visible to
  cause me to change my mind away from the estimate derived from a uniform
  prior on such features?}
\end{quote}
For an action that yields a new view, $j$, we compute $\delta'$ as follows
\begin{equation*}
\argmax_{\mathbf{Z}}\hspace{-5ex}
\sum_{\substack{\mathbf{Z}_1\ldots\mathbf{Z}_n\;\mathbf{Z}_j\\
\Phi[\mathbf{Z},\mathbf{Z}_1]\wedge\ldots\wedge\Phi[\mathbf{Z},\mathbf{Z}_n]
\wedge\Phi[\mathbf{Z},\mathbf{Z}_j]}}\hspace{-6ex}
\Pr(\mathbf{Z},\mathbf{Z}_1,\ldots,\mathbf{Z}_n,\mathbf{Z}_j)=\mathbf{Z}
\end{equation*}
when:
\begin{equation*}
\begin{array}{l}
\Pr(Z_{iq}^{\,f}=\neg\hat{Z}_{iq}^{\,f})=\frac{1}{2}+\delta\\
\Pr(Z_{iq}^{\,f}=\hat{Z}_{iq}^{\,f})=\frac{1}{2}-\delta
\end{array}
\end{equation*}
for all $q^{\,f}$ where
$V_{jq}^{f}=\true\wedge(\forall i)V_{iq}^{\,f}=\false$.
Because we wish to select the action with the smallest~$\delta'$, we need its
actual value.
Thus we perform binary search to find~$\delta'$ for each hypothetical action
and select the one with the lowest~$\delta'$.
This nominally requires sufficiently deep binary search to compute~$\delta'$ to
arbitrary precision.
One can make this process even faster by performing binary search on all
hypothetical actions simultaneously and terminating when there is only one
hypothetical action lower than the branch point.
This requires that binary search be only sufficiently deep to discriminate
between the available actions.

\section{Natural language}
\label{sec:language}

An interesting feature of our framework is that it allows for elegant inclusion
of information from other modalities.
Natural language, for example, can be integrated into our approach to draw
additional evidence for structure estimation from utterances describing the
structure in question.
A sentence, or set of sentences, describing a structure need not specify the
structure unambiguously.
Much like additional images from novel viewpoints can provide supplementary but
partial evidence for structure estimation, sentences providing incomplete
descriptions of structural features also can provide supplementary but partial
evidence for structure estimation.

We have investigated this possibility via a small domain-specific language for
describing some common features present in assembly toys.
This language has: two nouns (\emph{window} and \emph{door}), four spatial
relations (\emph{left of}, \emph{right of}, \emph{perpendicular to}, and
\emph{coplanar to}), and one conjunction (\emph{and}).
Sentences constructed from these words can easily be parsed into logical
formulas.

Analogous to how a CSP encodes the validity of an assembly through a set of
constraints, such logical formulas derived from sentential descriptions can also
constrain the structures to be considered.
The words in our vocabulary impose the following constraints:
\begin{compactenum}
\item A \emph{door} is composed of a rectangular vertical coplanar set of grid
  points.
  All grid points inside the door must be unoccupied.
  All grid points on the door posts must be log ends facing away from the door.
  All grid points on the mantel must be occupied by the same log.
  The threshold must be unoccupied and at the bottom of the structure.
\item A \emph{window} is similar to a door whose threshold is occupied by the
  same log and is not constrained to be at the bottom of the structure.
\item \emph{Perpendicular to} constrains the grid points of two
  entities to lie on perpendicular axes.
  \emph{Coplanar to} is analogous.
\item \emph{Right of} or \emph{left of} constrain the relative coordinates of
  the grid points of two entities
\end{compactenum}

We compute a joint multiple-view and natural-language structure estimate as
follows.
Let~$\mathbf{Z}$ be a collection of random variables~$Z_q$ associated with log
occupancy for a given initial structure.
Given a set of constraints $\Psi$ derived from natural language
and multiple views $i=1,\ldots,n$ with collections~$\mathbf{Z}_i$ of random
variables~$Z^+_q$, $Z^-_q$, $Z^u_q$, $Z^v_q$, and~$Z^w_q$ associated with the
image evidence for log features from those views, we compute
\begin{equation*}
\argmax_{\mathbf{Z}}\hspace{-12pt}
\sum_{\substack{\mathbf{Z}_1\ldots\mathbf{Z}_n\\
    \Phi\left[\mathbf{Z},\mathbf{Z}_1\right]\wedge\ldots\wedge\left[\mathbf{Z},\mathbf{Z}_n\right]\wedge\Psi\left[\mathbf{Z}\right]}}
\hspace{-10pt}\Pr\left(\mathbf{Z},\mathbf{Z}_1,\ldots,\mathbf{Z}_n\right)
\end{equation*}
An example of how such an extension improves results is shown in
Fig.~\ref{fig:language}.

\section{Results}
\label{sec:results}

We gathered a corpus of~5 different images of each of~32 different structures,
each from a different viewpoint, for a total of~160 images.
The structures were carefully designed so that proper subset relations exist
among various pairs of the~32 distinct structures.
Our video supplement demonstrates additional examples of our method.

\begin{figure}[t]
\begin{center}
\includegraphics[width=0.48\textwidth]{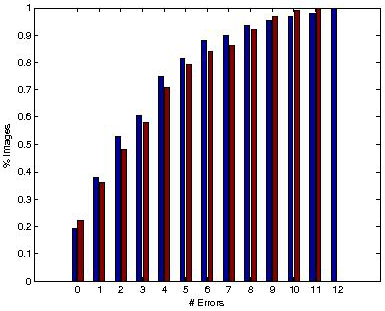}
\end{center}
\caption{Error histograms for manual visibility annotation (in blue) and
  automatic visibility estimation (in red).
100\% of the structures estimated had 12 or fewer errors.
Note that the latter performs as well as the former.}
\label{fig:visibility}
\end{figure}

We first evaluated automatic visibility estimation.
We performed combined visibility and structure estimation on~105 of the~160
images and compared the maximum-likelihood structure estimate to that produced by
\cite{Siddharth2011} using manual annotation of visibility.
For each image, we compare the maximum-likelihood structure estimate to ground
truth and compute the number of errors.
We do this as follows.
Each 1-, 2-, or 3-notch log in either the ground truth or estimated structure
that is replaced with a different, possibly empty, collection of logs in the
alternate structure counts as a single error (which may be a deletion,
addition, or substitution).
Further, each collection of~$r$ adjacent logs with the same medial axis in
the ground truth that is replaced with a different collection of~$s$ logs in
the estimated structure counts as $min(r,s)$ errors.
We then compute an error histogram of the number of images with fewer than~$t$
errors.
Fig.~\ref{fig:visibility} shows the error histograms for manual visibility
annotation and automatic visibility estimation.
Note that the latter performs as well as the former.
Thus our automatic visibility-estimation process appears to be reliable.

\begin{figure}[t]
\begin{center}
\includegraphics[width=0.48\textwidth]{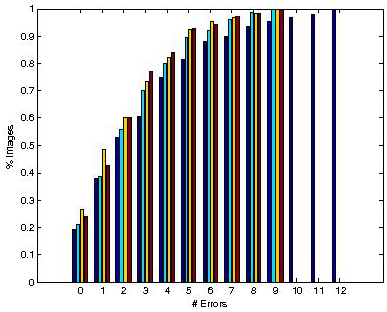}
\end{center}
\caption{Error histograms for the baseline structure estimation (in dark blue)
  and each of the active-vision process (partial disassembly in light blue,
  multiple views in yellow, and the combination of these in red).}
\label{fig:active}
\end{figure}

We then evaluated structure-estimation confidence assessment.
We computed the false-positive rate and false-negative rate of our
confidence-assessment procedure over the entire corpus of~105 images, where a
false positive occurs with a positive confidence assessment for an
incorrect structure estimate and a false negative occurs with negative
confidence assessment for a correct structure estimate.
This resulted in only~3 false positives and~7 false negatives on our corpus.

\begin{figure*}[t]
\begin{center}
\begin{tabular}{@{}c@{\hspace*{2pt}}c@{\hspace*{2pt}}c@{\hspace*{2pt}}c@{}}
\includegraphics[width=0.243\textwidth]{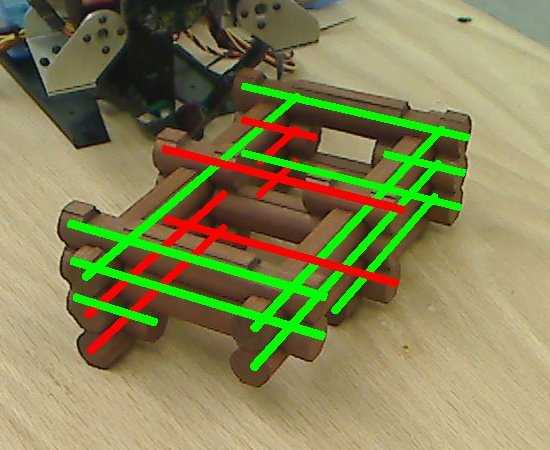}&
\includegraphics[width=0.243\textwidth]{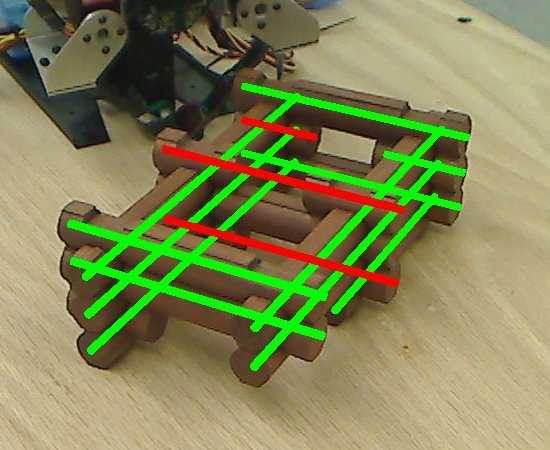}&
\includegraphics[width=0.243\textwidth]{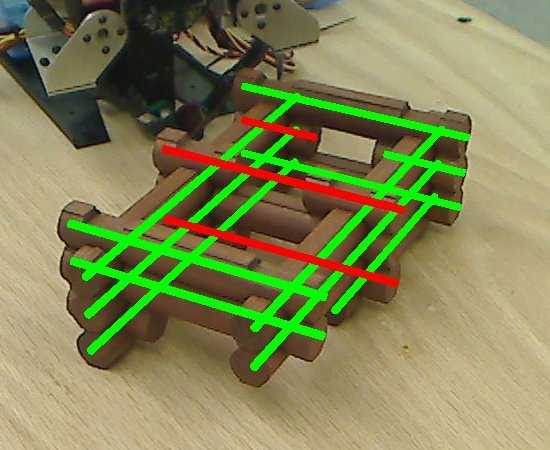}&
\includegraphics[width=0.243\textwidth]{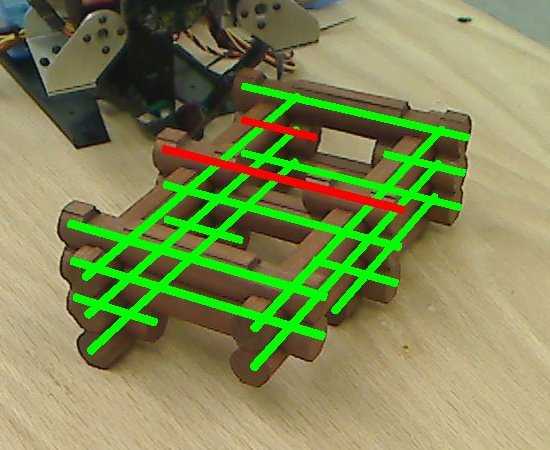}\\
(a)&(b)&(c)&(d)
\end{tabular}
\end{center}
\caption{\small Rendered structure for the following four methods:
(a)~Baseline structure estimation.
(b)~Partial disassembly.
(c)~Multiple views.
(d)~Combined partial disassembly and multiple views.}
\label{fig:results}
\end{figure*}

We then evaluated the active-vision process for performing actions to improve
structure-estimation confidence on~90 images from our
corpus.\footnotemark[\value{footnote}]
So as not to render this evaluation dependent on the mechanical reliability of
our robot which is tangential to the current paper and focus the evaluation on
the computational method, we use the fact that our corpus contains multiple
views of each structure from different viewpoints to simulate moving the robot
head to gather new views and the fact our corpus contains pairs of structures
in a proper-subset relation to simulate using the robot to perform partial
disassembly.
We first evaluated simulated robot-head motion to gather new views.
For each image, we took the other images of the same structure from different
viewpoints as potential actions and perform our active-vision process.
We next evaluated simulated robotic disassembly.
For each image, we took images of proper-subset structures taken from the same
viewpoint as potential actions and perform our active-vision process.
We finally evaluated simulated combined robot-head motion and robotic
disassembly.
For each image, we took all images of proper-subset structures taken from any
viewpoint as potential actions and perform our active-vision process.
For each of these, we computed the error histogram at the termination of the
active-vision process.
Fig.~\ref{fig:active} shows the error histograms for each of the active-vision
processes together with the error histogram for baseline structure estimation
from a single view on this subset of~90 images.
Fig.~\ref{fig:results} shows a rendering of the final estimated structure
when performing each of the four processes from Fig.~\ref{fig:active} on the
same initial image.
Log color indicates correct (green) or incorrect (red) estimation of log
occupancies.
Note that our active-vision processes consistently reduce estimation error.

\begin{figure*}
\begin{center}
\begin{tabular}{@{}c@{\hspace*{2pt}}c@{\hspace*{2pt}}c@{\hspace*{2pt}}c@{}}
\includegraphics[width=0.243\textwidth]{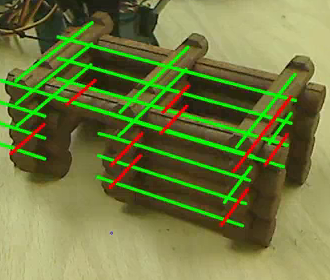}&
\includegraphics[width=0.243\textwidth]{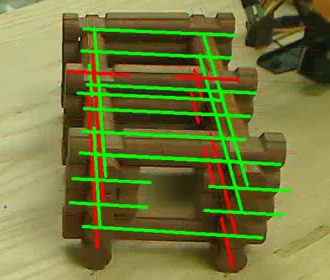}&
\includegraphics[width=0.243\textwidth]{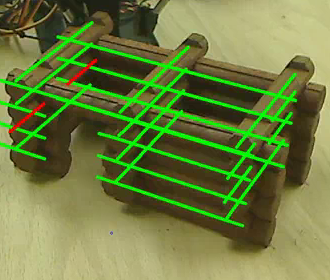}&
\includegraphics[width=0.243\textwidth]{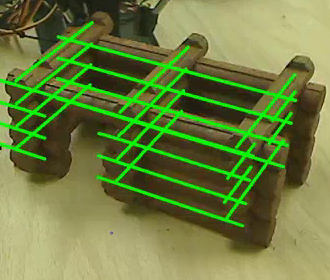}\\
(a)&(b)&(c)&(d)
\end{tabular}
\end{center}
\caption{\small An example of joint structure estimation from image evidence and natural language:
(a)~Baseline structure estimation from one view.
(b)~Structure estimation from a second view alone.
(c)~Structure estimation using information from both view from the viewpoint of the first view.
(d)~Structure estimation integrating the image evidence from the first view
with the sentence \emph{window left of and perpendicular to door}.}
\label{fig:language}
\end{figure*}

We demonstrate natural-language integration in
Fig.~\ref{fig:language}.
In Fig.~\ref{fig:language}(a), structure estimation is performed on a single
view, which due to occlusion, is unable to determine the correct structure.
A second view is acquired.
Note that this second view suffers from far more occlusion than the first view
and by itself produces a far worse structure (Fig.~\ref{fig:language}b) than the
first view alone.
The information available in these two views is integrated and jointly produces
a better structure estimate than either view by itself
(Fig.~\ref{fig:language}c).
However, this estimate is still imperfect.
To demonstrate the utility and power of integrating visual and linguistic
information, we intentionally discard the second view and construct a structure
estimate from just a single image together with a single linguistic
description, each of which is ambiguous taken in isolation.
The user provides the sentence \emph{window left of and perpendicular to door}.
Note this this sentence does not fully describe the assembly.
It does not specify the number of windows and doors, their absolute
positions, or the contents of the rest of the structure.
Yet this sentence together with the single image from the first view is
sufficient to correctly estimate the structure (Fig.~\ref{fig:language}d).

\section{Related work}
\label{sec:related}

{\noindent}Our work shares three overall objectives with prior work:
estimating 3D structure from 2D images, determining when there is
occlusion, and active vision.
However, our work explores each of these issues from a novel perspective.

Prior work on structure estimation (e.g. \citep{Saxena2007,Lee2009,Gupta2010})
focuses on \emph{surface estimation}, recovering a 3D surface from 2D images.
In contrast, our work focuses on recovering the \emph{constituent structure}
of an assembly: what parts are used to make the assembly and how such parts are
combined.
Existing state-of-the-art surface reconstruction methods (e.g. Make3D
\citep{Saxena2008}) are unable to determine surface structure of the kinds of
\LincolnLog\ assemblies considered here.
Ever if such surface estimates were successful, such estimates alone are
insufficient to determine the constituent structure.

Prior work on occlusion determination (e.g. \citep{Gupta2010,Hoiem2011})
focuses on finding occlusion boundaries: the 2D image boundaries of occluded
regions.
In contrast, our work focuses on determining occluded \emph{parts} in the
constituent structure.
We see no easy way to determine occluded parts from occlusion boundaries
because such boundaries alone are insufficient to determine even the number of
occluded parts, let alone their types and positions in a 3D structure.

Prior work on active vision (e.g. \citep{Maver1993}) focuses on integrating
multiple views into surface estimation and selecting new viewpoints to
facilitate such in the presence of occlusion.
In contrast, our work focuses on determining the confidence of constituent
structure estimates and choosing an action with maximal anticipated increase in
confidence.
We consider not only viewpoint changes but also robotic disassembly to view
object interiors.
Also note that the confidence estimates used in our approach to active vision
are mediated by the visual language model.
We might not need to perform active vision to observe all occluded structure as
it might be possible to infer part of the occluded structure.
Prior work selects a new viewpoint to render occluded structure visible.
We instead select an action to maximally increase confidence.
Such an action might actually not attempt to view an occluded portion of the
structure but rather increase confidence in a visible portion of the structure
in a way that when mediated by the language model ultimately yields a maximal
increase in the confidence assessment of a portion of the structure that
remains occluded even with the action taken.

\section{Conclusion}
\label{sec:conclusion}

We have presented a general framework for (a)~seeing the unseeable, (b)~seeing
unseeability, (c)~a rational basis for determining confidence in what one sees,
and (d)~an active-vision decision-making process for determining rational
behavior in the presence of unseeability.
We instantiated and evaluated our general framework in the
\LincolnLog\ domain and found it to be effective.
This framework has many potential extensions.
One can construct random variables to represent uncertain evidence in other
modalities such as language and speech and one can augment the stochastic CSP
to mutually constraint these variables together with the current random
variables that represent image evidence and latent structure so that a latent
utterance describes a latent structure.
One can then use the same maximum-likelihood estimation techniques to produce
the maximum-likelihood utterance consistent with a structure marginalizing
over image evidence.
This constitutes producing an utterance that describes a visual observation.
One can use the same maximum-likelihood estimation techniques to produce the
maximum-likelihood sequence of robotic actions consistent with building a
structure marginalizing over utterance evidence or alternatively image evidence.
This would constitute building a structure by understanding a linguistic
description of that structure or by copying a visually observed assembly.
One can combine evidence from an uncertain visual perception of a structure with
evidence from an uncertain linguistic description of that structure to reduce
the uncertainty of structure estimation.
This would constitute using vision and language to mutually disambiguate each
other.
One could augment one's collection of potential actions to include speech acts
as well as robotic-manipulation actions and search for the action that best
improves confidence.
This would constitute choosing between asking someone to provide you information
and seeking that information yourself.
One could determine what another agent can see from what that agent says.
Likewise one could decide what to say so that another agent can see what is
unseeable to that agent yet is seeable to you.
Overall, this can lead to a rational basis for cooperative agent behavior and a
theory of the perception-cognition-action loop which incorporates mutual
belief, goals, and desires where agents seek to assist each other by seeing
what their peers cannot, describing such sight, and inferring what their peers
can and cannot see.
We are currently beginning to investigate potential extensions to our
general approach and hope to present them in the future.

\section*{Acknowledgments}

This work was supported, in part, by NSF grant CCF-0438806, by the Naval
Research Laboratory under Contract Number N00173-10-1-G023, by the Army
Research Laboratory accomplished under Cooperative Agreement Number
W911NF-10-2-0060, and by computational resources provided by Information
Technology at Purdue through its Rosen Center for Advanced Computing.
Any views, opinions, findings, conclusions, or recommendations contained or
expressed in this document or material are those of the author(s) and do not
necessarily reflect or represent the views or official policies, either
expressed or implied, of NSF, the Naval Research Laboratory, the Office of
Naval Research, the Army Research Laboratory, or the U.S. Government.
The U.S. Government is authorized to reproduce and distribute reprints for
Government purposes, notwithstanding any copyright notation herein.

\nocite{Zach2008}
\nocite{Roy2004}
\nocite{Delage2006}
\nocite{Mackworth1977}
\nocite{Biederman1987}
\nocite{Fischler1981}
\nocite{Paletta2000}
\nocite{Tarabanis1995}
\nocite{Savova2009}
\nocite{Lippow2008}
\nocite{Zhu2006}
\nocite{Zhu2007}
\nocite{Heitz2008}

\bibliographystyle{plainnat}
\bibliography{arxiv2012c}

\end{document}